%% file: main.tex
\documentclass{vgtc}                          

\ifpdf
  \pdfoutput=1\relax                   
  \usepackage{graphicx}                
  \DeclareGraphicsExtensions{.pdf,.png,.jpg,.jpeg} 
\else
  \usepackage{graphicx}                
  \DeclareGraphicsExtensions{.eps}     
\fi%

\graphicspath{{figures/}{pictures/}{images/}{./}} 

\usepackage{microtype}                 
\PassOptionsToPackage{warn}{textcomp}  
\usepackage{textcomp}                  
\usepackage{mathptmx}                  
\usepackage{times}                     
\usepackage{cite}                      
\usepackage{tabu}                      
\usepackage{booktabs}                  
\usepackage{enumitem}
\usepackage{subfig}

\newcommand{\tool}[1]{LinguisticLens}

\vgtccategory{Research}

\vgtcinsertpkg

\title{Visualizing Linguistic Diversity of Text Datasets \\Synthesized by Large Language Models}

\author{Emily Reif %
\and Minsuk Kahng%
\and Savvas Petridis\thanks{
e-mails: {ereif$|$kahng$|$petridis}@google.com}}

\affiliation{\scriptsize Google Research}

\teaser{\input{05-teaserfigure}
 }
\abstract{\input{00-abstract}
} 

\CCScatlist{
  \CCScatTwelve{Visualization}{Text}{MLStatsModel}
}

\begin{document}



\maketitle
\input{10-introduction}

\input{20-background}

\input{30-relatedwork}

\input{40-challenges}

\input{50-visualization}

\input{60-casestudies}

\input{80-discussions}

\acknowledgments{We thank Ellen Jiang, Martin Wattenberg, Ann Yuan, Lucas Dixon, and Google People+AI Research (PAIR) team for their feedback.}


\bibliographystyle{abbrv-doi-hyperref}

\bibliography{references}
\end{document}

%% file: 05-teaserfigure.tex
\centering
  \includegraphics[width=\linewidth]{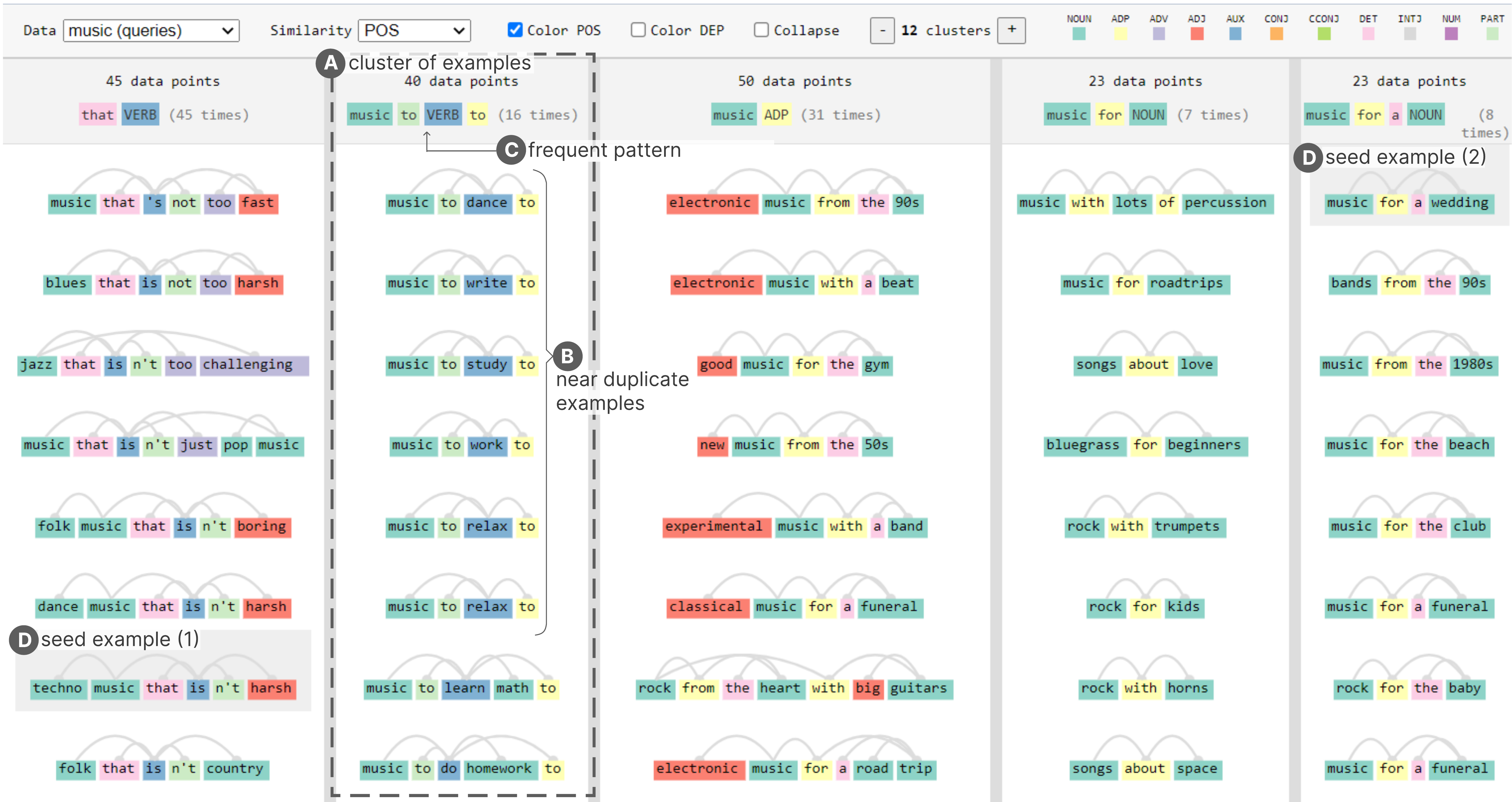}
  \caption{\tool{}, a new visualization tool for making sense of text datasets synthesized by large language models (LLMs) and analyzing the diversity of examples. \textbf{(A)} Each column represents a cluster of examples, where clustering is performed based on their syntax, tokens, or embeddings. Each example within the column is colored by part-of-speech (POS) tag, and has the dependency parse tree in gray.
  \textbf{(B)} In this example, users can easily find a group of examples very similar to each other.
  \textbf{(C)} Each cluster has a summary string, showing one of the most frequent subpattern across the examples. 
  These text examples are generated with \textit{few-shot prompting} on LLMs with \textbf{(D)} some seed examples. 
  }
  \label{fig:teaser}

%% file: 00-abstract.tex
Large language models (LLMs) can be used to generate smaller, more refined datasets via few-shot prompting for benchmarking, fine-tuning or other use cases. However, understanding and evaluating these datasets is difficult, and the failure modes of LLM-generated data are still not well understood. Specifically, the data can be repetitive in surprising ways, not only semantically but also syntactically and lexically.
We present \tool{}, a novel interactive visualization tool for making sense of and analyzing syntactic diversity of LLM-generated datasets. \tool{} clusters text along syntactic, lexical, and semantic axes. It supports hierarchical visualization of a text dataset, allowing users to quickly scan for an overview and inspect individual examples. The live demo is available at \url{https://shorturl.at/zHOUV}.

%% file: 10-introduction.tex
\section{Introduction}

Large language models (LLMs) are becoming ubiquitous for their ability to solve a wide range of linguistic tasks with \textit{prompting} that does not require additional model training~\cite{DBLP:journals/corr/abs-2005-14165, workshop2023bloom, anil2023palm}.   
This ability also lets them generate smaller, more refined datasets for finetuning~\cite{tang2023does, vijayakumar2022evaluating, synthetic_patient_data}, 
benchmarking~\cite{DBLP:journals/corr/abs-2111-06467}, low-resource tasks or languages~\cite{he2022synthetic, borisov2022language}, and counterfactual testing (e.g., examples that are identical except for having different religious or gender-based identities~\cite{fryer2022flexible}).

A critical challenge lies in making sense of these generated datasets and evaluating their quality. Given that the desired tasks are often novel and have no existing dataset or ground truth by definition, automatically evaluating the quality of these generated examples with certain metrics is not straightforward. Although crowd workers can evaluate the quality of individual examples, it is costly, and finding patterns across large amounts of text examples remains a challenge.
Moreover, understanding the specific failure modes of LLMs is still an evolving area, and these undesirable generated output trends can be hard to spot. In particular, generated examples often overfit to the seed examples in unexpected ways. One such pathology is syntactic overfitting, where generated examples are grammatically similarly or identical to the seed data. This can be difficult to find as a single overfit example is not a problem, but if larger portions of the dataset has the same syntactic structure, it is a significant issue for dataset diversity. The same is true for lexical overfitting, where specific words appear frequently in the generated dataset more often than is desired.

In this paper, we present \tool{}, a novel interactive visualization tool for making sense of synthetically-generated text datasets. \tool{} specifically focuses on analyzing the syntactic and lexical diversity of datasets. It allows users to explore groups of examples that are clustered based on their syntactic structure and lexical overlap.
Clusters can be based on other text similarity methods too, including embedding similarity, and we find that our approach is more effective for analyzing the diversity of synthetic datasets.

\tool{} runs on web browsers and users simply need to provide their datasets as CSV files. A live demo can be found at \url{https://shorturl.at/zHOUV}. The source code is available at \url{https://github.com/PAIR-code/interpretability}.

%% file: 20-background.tex
\section{Background: Synthesizing Datasets using LLMs}
\label{background}

This section provides a brief background about how people generate datasets using LLMs. Suppose we want to create a small dataset of music recommendations to fine tune a music recommendation model that returns a set of artists based on a short query provided by a user.
An example datapoint might have a query, \texttt{`oldies but goodies'}, and a label \texttt{`Aretha Franklin, The Beach Boys, Stevie Wonder, The Supremes, Bill Withers.'} 

Within the prompt, we can provide a few examples that the model can use to generate similar datapoints.
This is known as \textit{few-shot}, or \textit{in-context}, \textit{learning}. For example, the prompt could be:

\begin{verbatim}
   Query: {oldies but goodies}
   Recommended Artists: {Aretha Franklin, Madonna}
   Query: {music that makes you want to dance}
   Recommended Artists: {Kraftwerk, The Cure, B-52s}
   Query: {
\end{verbatim}
The model will continue the text following this pattern (e.g., see below), and from this we can parse a new set of examples. With this approach, LLMs can create hundreds or thousands of these synthetic examples. Our goal is to make sense of them.

\begin{verbatim}
   chill out music}
   Recommended Artists: {Bonobo, Massive Attack}
   Query: {female vocalists}
   Recommended Artists: {Carole King, Joni Mitchell}
\end{verbatim}

%% file: 30-relatedwork.tex
\section{Related work}
\subsection{Evaluating Datasets Generated by LLMs}
Evaluating LLM-generated datasets is not a straightforward task.
In the best case, one can measure downstream performance of a model trained on that data~\cite{tang2023does, borisov2022language}. When this is impossible (e.g., benchmarks or a new task), the dataset quality must be evaluated with defined metrics~\cite{synthetic_patient_data}.
Automatic methods for evaluating LLM-generated datasets include measuring the similarity between the original distribution and generated distribution~\cite{tang2023does}, but this is difficult for unstructured text, and presupposes that a golden corpus exists, which would obviate the need for synthetic data in the first place. Yuan et al.~\cite{DBLP:journals/corr/abs-2111-06467} use human evaluation to determine the quality of individual examples. Lara et al.~\cite{lara2022evaluation} measure diversity along annotated features, such as topics and sentiment. Other methods involve visualization-based human-in-the-loop qualitative evaluation. In the image domain, Assogba et al.~\cite{assogba2023large} built a visualization tool with embedding-based clusters for comparing the distributions of generated and ground truth images. 

\subsection{Visualizing Text Corpora}
There is a large body of visualization research
on making sense of large text corpora~\cite{corpus, doi:10.1057/palgrave.ivs.9500180, 4658133}. 
Clustering, including topic modelings, has been a popular approach to organizing text datasets~\cite{kucher2015text, 5613456, choo2013utopian}.
Visualization is also a ubiquitous tool for machine learning interpretability~\cite{hohman2018visual, yuan2021survey, chatzimparmpas2020survey}, including those for text data~\cite{strobelt2017lstmvis, tenney2020language, reif2019visualizing}.
Reif et al.~\cite{reif2019visualizing} investigated how a pretrained BERT model internally represents syntactic information through a range of visualizations.
However, visual analysis of text datasets synthesized from LLMs, especially their syntactic structures, is underexplored~\cite{brath2023role, collins2022visual}.

%% file: 40-challenges.tex
\section{Design Challenges}
We spoke to several software engineers and researchers at Google who are using LLMs to synthesize data about the challenges they had evaluating and understanding it. 
These use cases included music queries, adversarial examples for fairness analysis benchmarks, medical questions, and to-do lists (with no personally identifiable information), among others. 
Below we summarize our key findings.

\begin{enumerate}[label=C\arabic*.]
    
\item \textbf{Quick overview of datasets.}
Most of the practitioners we spoke to first wanted to get a quick overview of the full dataset. This included understanding what basic examples look like, how much these examples tended to differ from each other, and how much the examples matched the seed data.

\item \textbf{Identifying groups of examples and seeing their distributions.}
Practitioners also discussed trying to finding patterns, specifically identifying groups of examples that share common characteristics, and seeing their distributions. 
For example, in the case of to-do lists, there were patterns in how long each of the list items were, and how many items there were overall.

\item \textbf{Finding near duplicates.}
Examples are sometimes too similar to each other: one of the most common challenges we discussed was finding near duplicates. For example, many music recommendation queries follow the form \texttt{`[BLANK] that [BLANK] like [BLANK]'} (e.g., `music that sounds like nature').
There are many almost identical examples where a single word is swapped out. These are difficult to spot, e.g., when the words are mostly different, but the overall phrasing is identical across a set of examples, when manually scanning a long unordered list of data points.
People found this a particularly thorny problem to address. Exact duplicates are usually unequivocally bad and easy to find. However, what constituted an undesirable \textbf{near} duplicate was not only dependent on use cases, but also difficult to automatically detect.

\item \textbf{Understanding dataset diversity.}
Practitioners also wanted to know whether or not the dataset was diverse. However, like near duplication, the notion of diversity is hard to pin down and differs depending on the use case: for the to-do lists, it means having a broad range of words and topics in the lists; for the music queries, it means having a range of different phrasings, and having outliers that were far from the norm but still relevant; and for the medical questions dataset, it is important that the questions are phrased in different ways. 

\end{enumerate}

The overall consensus from the practitioners was that synthesizing data with LLMs was a novel task, without a robust evaluation method or sensemaking tools to support it. Existing tools, like topic modeling, could support some of the challenges (especially \textbf{C1} and \textbf{C2}), but focus on individual tokens, rather than longer sequences or other structural properties of the text.  Thus, we decided to design a new tool that addresses the challenges that we identified.

%% file: 50-visualization.tex
\section{Visualization Design of \tool{}}

We designed and developed \tool{} for exploring and evaluating the linguistic diversity of synthesized text datasets. \tool{} specifically focuses on finding patterns and fuzzy duplicates across a set of examples. The details of our design are as follows.

\subsection{Introducing \tool{}}

\autoref{fig:teaser} shows an interface of \tool{}, which uses Lit\footnote{\url{https://lit.dev/}}, a declarative web framework, and D3\footnote{\url{https://d3js.org/}} for visualizing each example as a graph structure.
It consists of multiple columns (six in this figure) each representing a cluster of examples. Each cluster lists text examples vertically (e.g., about 12 examples in this figure).
Each text example (e.g., query text for a music recommender) is represented as a sequence of tokens, shown as a colored rectangle.
We first describes how we visualize each example and then how we cluster and visualize them for exploration and analysis.

\begin{figure}[!bt]
 \centering 
 \includegraphics[width=0.9\columnwidth]{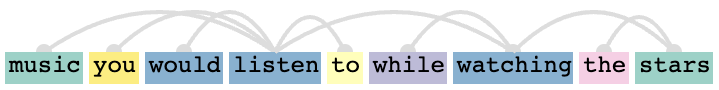}
 \vspace{-5pt}
 \caption{Visualization of a single example in \tool{}. Words are colored by their parts of speech, with the dependency graph shown as arcs in gray.}
 \label{fig:single_example}
\end{figure}

\subsection{Visualizing Individual Examples}

As illustrated in \autoref{fig:single_example},
each example (e.g., a query) is visually represented as a sequence of tokens, where each token is colored based on its part-of-speech (POS) tag. The dependencies among the tokens, extracted from the dependency parse graph is shown as arcs.

\begin{figure}[!bt]
 \centering 
 \includegraphics[width=\columnwidth]{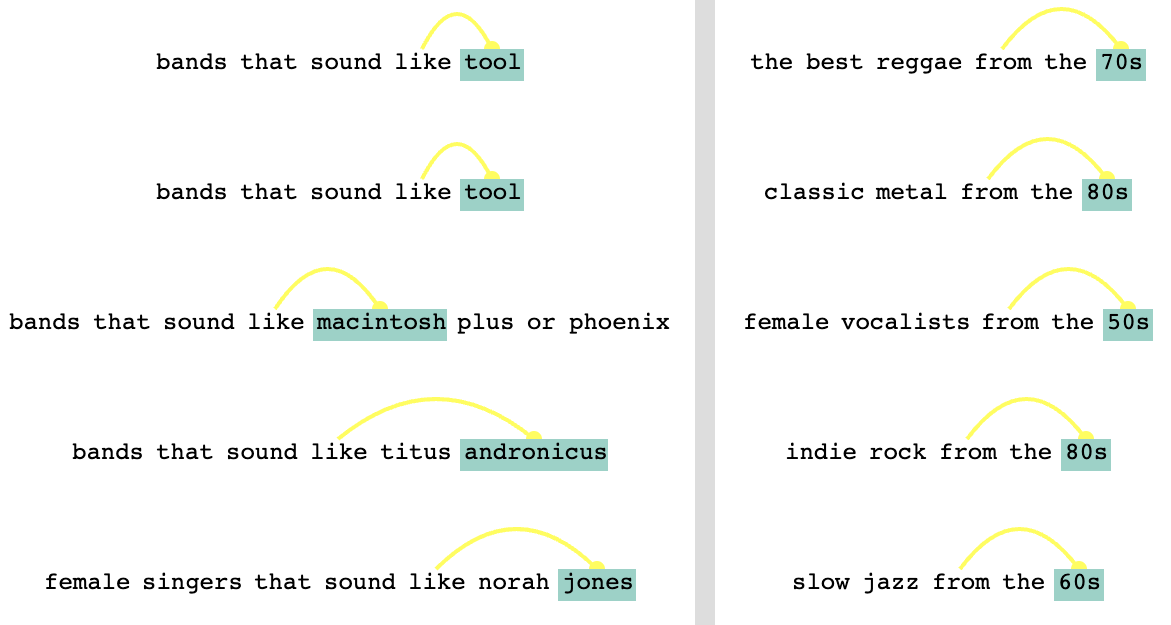}
 \vspace{-5pt}
 \caption{When a user hovers on a token, we highlight other tokens with the same POS tag and dependency relationship. }
 \label{fig:hover}
\end{figure}

\tool{} displays multiple examples in a vertical column and provides users with interactions to easily identify patterns across the examples. See section \autoref{sec:clusters} for how these are clustered into columns.
For instance, when a user hovers over a token in an example, \tool{} highlights the hovered token and the dependency link to that token, as well as tokens and links from other examples with the same grammatical relationship (e.g., from \texttt{NOUN} to \texttt{ADP}). Since these grammatical features are precalculated, it is trivial to filter in this way. \autoref{fig:hover} shows an example when hovering over the token `\texttt{tool}'. This enables users to quickly discover repeated substructures of examples in the dataset.

\subsection{Visualizing Clusters of Examples}
\label{sec:clusters}
To help users easily discover linguistic patterns (near duplicates (\textbf{C3}), diversity (\textbf{C4})) across groups of multiple examples (\textbf{C2}),
we design \tool{} to cluster examples by multiple different metrics (e.g., POS, embedding, and token) and allow users to choose one. These fuzzy duplicate clusters are shown in columns, with some metadata about each cluster (e.g., count, frequent pattern). 

\subsubsection{Clustering with Syntax: Technical Details} 

\indent\indent
\textbf{Syntactic Feature Extraction.}
To represent each example with its syntactic information, we extract a number of features.
We first represent each example in one of the three ways: as a sequence of its individual tokens, as a sequence of the POS tags for those tokens, and as a sequence of grammatical dependencies for those tokens.\footnote{We used the spaCy library~\cite{spacy2} to extract POS and dependency.} We then populate their \textit{n}-grams ($n=1, 2, 3$) from one of these approaches. 
Then we define pair-wise similarity between two examples by computing the \textit{n}-gram overlaps and dividing by the number of tokens in longer text.
While embedding examples using pretrained models is a well-known approach to representing unstructured text datasets,
the embedding encodes both semantic and syntactic information. Thus, it is less effective for revealing syntactic and lexical diversity of generated datasets.

\textbf{Hierarchical Clustering.}
We run an agglomerative clustering algorithm based on the similarity metrics defined from the extracted features.
From a dendrogram tree returned by the algorithm, we form a list of flattened clusters for different numbers of clusters ($k=3, 5, 10, ..., 50$).
For each cluster, we maintain an ordered list of examples where its order follows the order of the leaves in the dendrogram tree, similar to the approach used in DendroMap~\cite{bertucci2022dendromap}.
This places similar examples next to each other, which can enable users to easily identify groups of similar examples, especially near-duplicates, when they are visualized.

\begin{figure}[!tb]
 \centering 
 \includegraphics[width=\columnwidth]{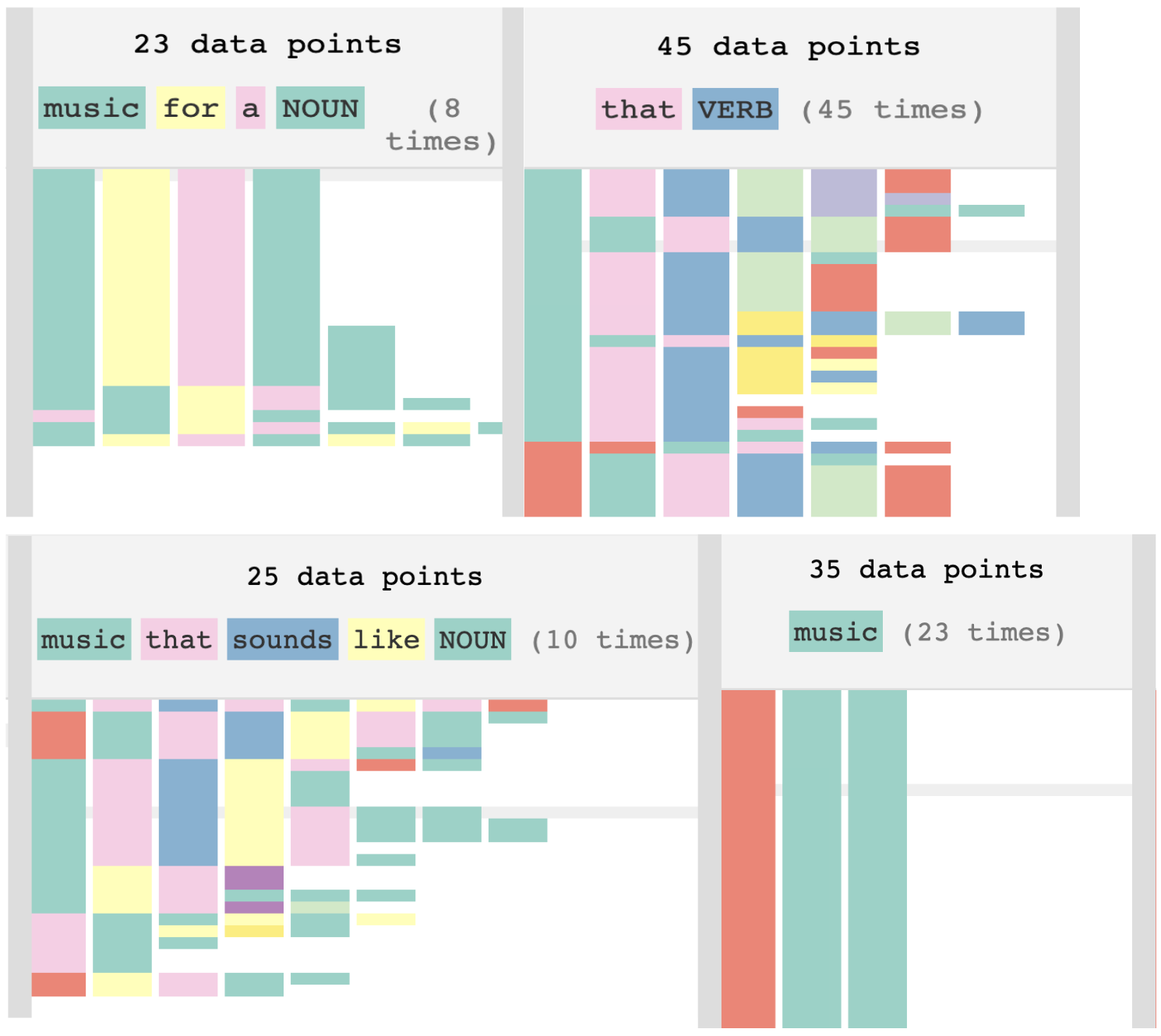}
 \caption{\tool{} supports the \textbf{collapsed} mode for overview. By adapting the idea of Table Lens, it displays a thumbprint of examples, enabling a quick overview of  their linguistic patterns and the distributions across clusters.}
 \label{fig:collapsed}
\end{figure}

\subsubsection{Overview by Adapting Table Lens}

\tool{} consists of multiple columns each representing one of the clusters.
An important challenge in visualizing text datasets is that we cannot concurrently display many examples in screen, and it would visually overwhelming even if we could.
To effectively provide an overview of large number of examples (\textbf{C1}),
we adapt Table Lens~\cite{tableLens}.
As shown in \autoref{fig:collapsed}, \tool{} collapses examples by default, but the user can expand them. It significantly saves screen space while still presenting the grammatical patterns of individual examples based on their POS tokens encoded by colors. This enables users to quickly identify a range of linguistic patterns and their distributions. For example, we see a cluster (top left) of repetative examples that follow the pattern \texttt{NOUN ADP DET NOUN}.

\begin{figure}[tb]
 \centering 
 \includegraphics[width=.6\columnwidth]{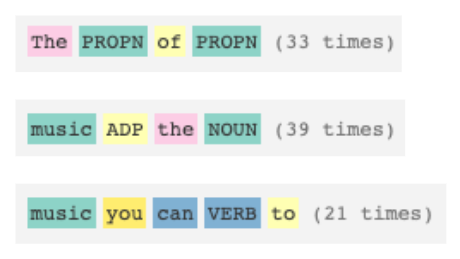}
 \vspace{-5pt}
 \caption{Each cluster is summarized with a frequent subsequence pattern determined based tokens and parts-of-speech of examples.
 }
 \label{fig:summaries}
\end{figure}

\subsubsection{Summary with Frequent Pattern Mining}
To further help users quickly get a sense of the type of examples for each cluster,
we display the most frequent sequential pattern at the top of each cluster.
The pattern is formed as a sequence of items where each item is either a text token or POS tag (e.g., (music, you, can, VERB, to) as shown in the last one in \autoref{fig:summaries}), where VERB is a POS tag, and the other four items are tokens.
We additionally display the number of occurrences of this pattern.

\textbf{Algorithm Details.} We extract the most frequent pattern by adapting a well-known sequential pattern mining algorithm~\cite{han2001prefixspan}.
To form inputs to the pattern mining algorithm, we transform each example into a sequence of token or its POS's (e.g., (music$\vert$NOUN, you$\vert$PRON, can$\vert$VERB, dance$\vert$VERB, to$\vert$ADP), allowing patterns that are a mix of actual tokens and POS's.
Once several candidate patterns are generated,
we rank them to determine the most representative.
We score each pattern with a heuristic linear function that incorporates a few signals, such as the number of occurrences (higher score if appearing more), their length (higher score if pattern has a larger number of tokens or POS's), and proportion of tokens/POS's (higher score if actual tokens are used more than POSes).
For instance, a pattern \texttt{(music, you, can, VERB, to)} can get a higher score than another pattern \texttt{(music, PRON, can, VERB)}, even if the latter appears more times, since the former is longer contains more actual tokens.
We return the highest scored pattern as the most informative pattern that summarizes each cluster.

%% file: 60-casestudies.tex
\section{Case studies}
\label{sec:casestudies}
We demonstrate \tool{} on two different use cases described below. These were not run with domain experts, but do use real data and were based on findings from our initial conversations. To generate the datasets, we started with 5-10 hand-crafted examples, expanding them to 500 examples using the method described in \autoref{background} with PaLM 2~\cite{anil2023palm}.
For data privacy reasons, we do not include all the use cases from the motivating user conversations.

\subsection{Dialog Agent}
Our first use case is expanding a few dialogue examples into a larger dataset for fine-tuning a LLM to be a dialogue model. Using \tool{}, we quickly see what individual examples look like (\textbf{C1}), and find some basic patterns (\textbf{C2}) from looking at clusters with different metrics. E.g., we see that most examples are of similar lengths (using the POS metric), that there is a variety of punctuation types (embedding metric), and that there is a group of examples asking for different kinds of favorites (token metric).

\begin{figure}[tb]
 \centering 
 \includegraphics[width=\columnwidth]{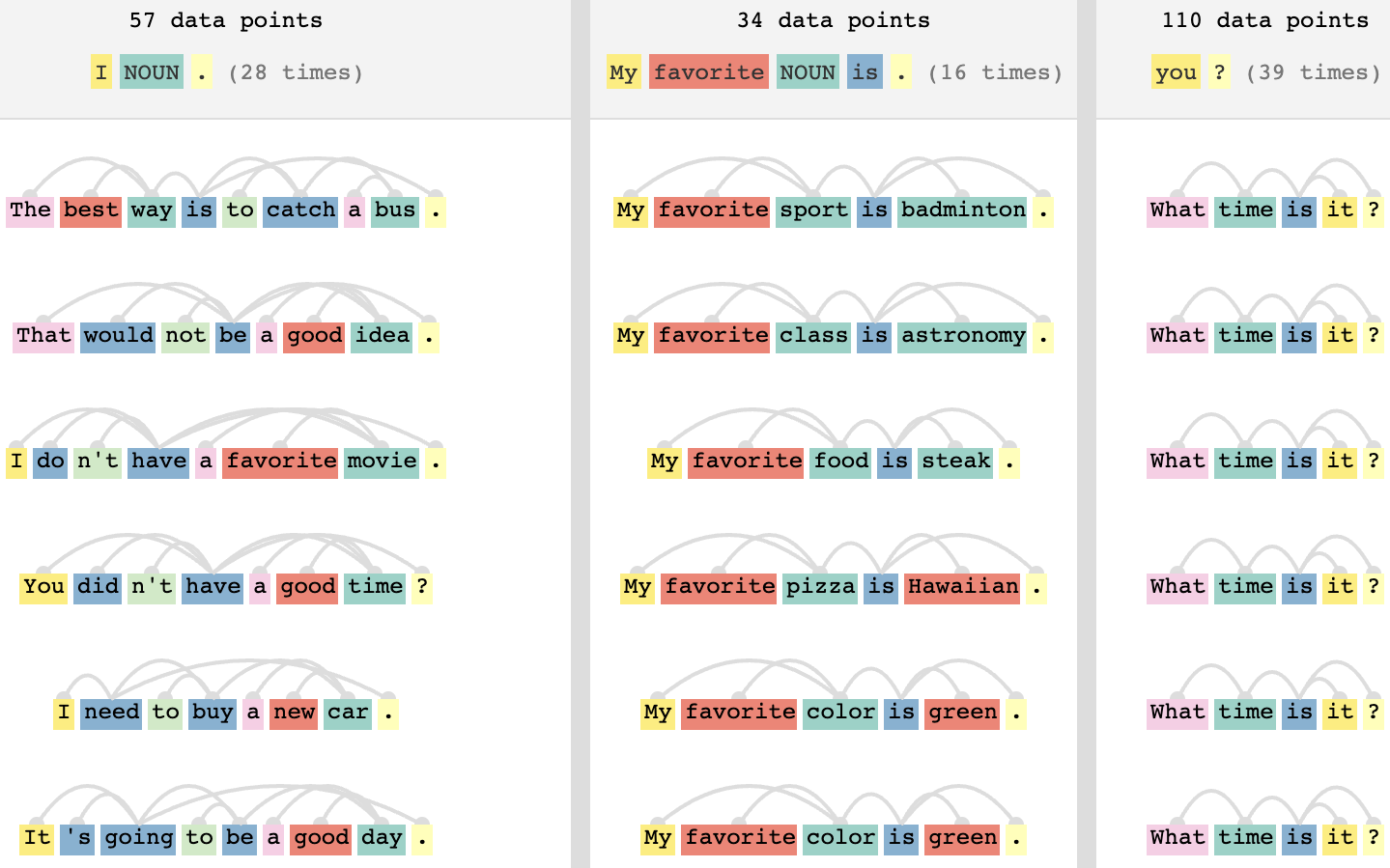}
 \vspace{-10pt}
 \caption{A dataset of dialog examples. Each cluster contains sets of syntactic near-duplicates.
 A seed example that is used in few-shot examples in prompt is shown with a gray box.
 }
 \label{fig:dialog_full}
\end{figure}

\textbf{Undesirable Repetition (C3).} We also see that there are repetitive examples in \autoref{fig:dialog_full}, for example, many instances of \texttt{`<I/we> <like/love> to...'} and \texttt{`what is your favorite...'} examples, which should be deduplicated for a more diverse dataset.

\textbf{Finding Outliers (C4).} Due to the nature of agglomerative clustering, the final cluster has outlier examples that are not similar enough to other examples to be added to a different cluster. This can be useful in two ways: sometimes these outliers are often either degenerate, or so far from the desired distribution that they should be deleted. However, they may also be beneficial, unique examples, which the user can incorporate into the prompt iteration.

\subsection{Music Recommendations}
Our second use case utilizes the music dataset described in  \autoref{background}. We explore an overview of examples in the dataset (\textbf{C1}) and look for some patterns in example phrasing and content (\textbf{C2}).

\textbf{Desirable Repetition.} Interestingly, we found that \tool{} was also useful for finding \textit{desirable} repetition, and pointed to areas of the dataset where more near-duplicates should be added. For example, in the music dataset, there is a cluster of \texttt{`best music from the 90s', `best music from the 70s'}, etc. These were actually deemed desirable: in fact, the user wanted to augment the dataset to include all decades. This points to the necessity of human in the loop evaluation: there is no way to automatically detect if duplication is good or bad in a given situation.

\textbf{Undesirable Repetition (C3).} As in the dialog case, we found clusters of examples with near-duplicate grammar (\autoref{fig:hover} - left column). These examples are individually interesting, but are syntactically so similar to each other the user was worried a downstream mode would overfit to this pattern of \texttt{`<nouns> that sound like <nouns>'}.

%% file: 80-discussions.tex
\section{Limitations and Future Work}
One area for future research is to design advanced human-in-the-loop approaches for users to interactively identifying example clusters based on metrics of interest. Another direction is to formalize the notion of syntactic overfitting and use it as an evaluation metric for different prompting strategies. A limitation of this work is scalability, both in terms of individual example lengths as well as number of examples. The current interface is optimized to examples that have less than 10 tokens, while the lengths of input and output for LLMs can be long.

%% file: main.bbl
\begin{thebibliography}{10}

\bibitem{anil2023palm}
\href{https://arxiv.org/abs/2305.10403}{R.~Anil, A.~M. Dai, O.~Firat,
  M.~Johnson, D.~Lepikhin, A.~Passos, S.~Shakeri, E.~Taropa, P.~Bailey,
  Z.~Chen, E.~Chu, J.~H. Clark, L.~E. Shafey, Y.~Huang, K.~Meier-Hellstern,
  G.~Mishra, E.~Moreira, M.~Omernick, K.~Robinson, S.~Ruder, Y.~Tay, K.~Xiao,
  Y.~Xu, Y.~Zhang, G.~H. Abrego, J.~Ahn, J.~Austin, P.~Barham, J.~Botha,
  J.~Bradbury, S.~Brahma, K.~Brooks, M.~Catasta, Y.~Cheng, C.~Cherry, C.~A.
  Choquette-Choo, A.~Chowdhery, C.~Crepy, S.~Dave, M.~Dehghani, S.~Dev,
  J.~Devlin, M.~Díaz, N.~Du, E.~Dyer, V.~Feinberg, F.~Feng, V.~Fienber,
  M.~Freitag, X.~Garcia, S.~Gehrmann, L.~Gonzalez, G.~Gur-Ari, S.~Hand,
  H.~Hashemi, L.~Hou, J.~Howland, A.~Hu, J.~Hui, J.~Hurwitz, M.~Isard,
  A.~Ittycheriah, M.~Jagielski, W.~Jia, K.~Kenealy, M.~Krikun, S.~Kudugunta,
  C.~Lan, K.~Lee, B.~Lee, E.~Li, M.~Li, W.~Li, Y.~Li, J.~Li, H.~Lim, H.~Lin,
  Z.~Liu, F.~Liu, M.~Maggioni, A.~Mahendru, J.~Maynez, V.~Misra, M.~Moussalem,
  Z.~Nado, J.~Nham, E.~Ni, A.~Nystrom, A.~Parrish, M.~Pellat, M.~Polacek,
  A.~Polozov, R.~Pope, S.~Qiao, E.~Reif, B.~Richter, P.~Riley, A.~C. Ros,
  A.~Roy, B.~Saeta, R.~Samuel, R.~Shelby, A.~Slone, D.~Smilkov, D.~R. So,
  D.~Sohn, S.~Tokumine, D.~Valter, V.~Vasudevan, K.~Vodrahalli, X.~Wang,
  P.~Wang, Z.~Wang, T.~Wang, J.~Wieting, Y.~Wu, K.~Xu, Y.~Xu, L.~Xue, P.~Yin,
  J.~Yu, Q.~Zhang, S.~Zheng, C.~Zheng, W.~Zhou, D.~Zhou, S.~Petrov, and Y.~Wu}.
\newblock \href{https://arxiv.org/abs/2305.10403}{{PaLM} 2 technical report}.
\newblock \href{https://arxiv.org/abs/2305.10403}{{\em arXiv preprint
  arXiv:2305.10403}}, \href{https://arxiv.org/abs/2305.10403}{2023}.

\bibitem{assogba2023large}
\href{https://arxiv.org/abs/2301.04518}{Y.~Assogba, A.~Pearce, and M.~Elliott}.
\newblock \href{https://arxiv.org/abs/2301.04518}{Large scale qualitative
  evaluation of generative image model outputs}.
\newblock \href{https://arxiv.org/abs/2301.04518}{{\em arXiv preprint
  arXiv:2301.04518}}, \href{https://arxiv.org/abs/2301.04518}{2023}.

\bibitem{bertucci2022dendromap}
\href{https://doi.org/10.1109/TVCG.2022.3209425}{D.~Bertucci, M.~M. Hamid,
  Y.~Anand, A.~Ruangrotsakun, D.~Tabatabai, M.~Perez, and M.~Kahng}.
\newblock \href{https://doi.org/10.1109/TVCG.2022.3209425}{{DendroMap}: Visual
  exploration of large-scale image datasets for machine learning with
  treemaps}.
\newblock \href{https://doi.org/10.1109/TVCG.2022.3209425}{{\em IEEE
  Transactions on Visualization and Computer Graphics}},
  \href{https://doi.org/10.1109/TVCG.2022.3209425}{29(1):320--330},
  \href{https://doi.org/10.1109/TVCG.2022.3209425}{2022}.
  \href{https://doi.org/10.1109/TVCG.2022.3209425}
{doi: {{%
10\hspace{.1pt}\discretionary{.}{%
}{.}\hspace{.4pt}1109\discretionary{/}{%
}{/}TVCG\hspace{.1pt}\discretionary{.}{%
}{.}\hspace{.4pt}2022\hspace{.1pt}\discretionary{.}{%
}{.}\hspace{.4pt}3209425}}}


\bibitem{borisov2022language}
\href{https://arxiv.org/abs/2210.06280}{V.~Borisov, K.~Seßler, T.~Leemann,
  M.~Pawelczyk, and G.~Kasneci}.
\newblock \href{https://arxiv.org/abs/2210.06280}{Language models are realistic
  tabular data generators}.
\newblock \href{https://arxiv.org/abs/2210.06280}{{\em arXiv preprint
  arXiv:2210.06280}}, \href{https://arxiv.org/abs/2210.06280}{2022}.

\bibitem{brath2023role}
R.~Brath, D.~Keim, J.~Knittel, S.~Pan, P.~Sommerauer, and H.~Strobelt.
\newblock The role of interactive visualization in explaining (large) {NLP}
  models: from data to inference.
\newblock {\em arXiv preprint arXiv:2301.04528}, 2023.

\bibitem{DBLP:journals/corr/abs-2005-14165}
T.~B. Brown, B.~Mann, N.~Ryder, M.~Subbiah, J.~Kaplan, P.~Dhariwal,
  A.~Neelakantan, P.~Shyam, G.~Sastry, A.~Askell, S.~Agarwal,
  A.~Herbert{-}Voss, G.~Krueger, T.~Henighan, R.~Child, A.~Ramesh, D.~M.
  Ziegler, J.~Wu, C.~Winter, C.~Hesse, M.~Chen, E.~Sigler, M.~Litwin, S.~Gray,
  B.~Chess, J.~Clark, C.~Berner, S.~McCandlish, A.~Radford, I.~Sutskever, and
  D.~Amodei.
\newblock Language models are few-shot learners.
\newblock {\em Advances in Neural Information Processing Systems},
  33:1877--1901, 2020.

\bibitem{5613456}
\href{https://doi.org/10.1109/TVCG.2010.154}{N.~Cao, J.~Sun, Y.-R. Lin,
  D.~Gotz, S.~Liu, and H.~Qu}.
\newblock \href{https://doi.org/10.1109/TVCG.2010.154}{{FacetAtlas}:
  Multifaceted visualization for rich text corpora}.
\newblock \href{https://doi.org/10.1109/TVCG.2010.154}{{\em IEEE Transactions
  on Visualization and Computer Graphics}},
  \href{https://doi.org/10.1109/TVCG.2010.154}{16(6):1172--1181},
  \href{https://doi.org/10.1109/TVCG.2010.154}{2010}.
  \href{https://doi.org/10.1109/TVCG.2010.154}
{doi: {{%
10\hspace{.1pt}\discretionary{.}{%
}{.}\hspace{.4pt}1109\discretionary{/}{%
}{/}TVCG\hspace{.1pt}\discretionary{.}{%
}{.}\hspace{.4pt}2010\hspace{.1pt}\discretionary{.}{%
}{.}\hspace{.4pt}154}}}


\bibitem{chatzimparmpas2020survey}
A.~Chatzimparmpas, R.~M. Martins, I.~Jusufi, and A.~Kerren.
\newblock A survey of surveys on the use of visualization for interpreting
  machine learning models.
\newblock {\em Information Visualization}, 19(3):207--233, 2020.

\bibitem{choo2013utopian}
\href{https://doi.org/10.1109/TVCG.2013.212}{J.~Choo, C.~Lee, C.~K. Reddy, and
  H.~Park}.
\newblock \href{https://doi.org/10.1109/TVCG.2013.212}{Utopian: User-driven
  topic modeling based on interactive nonnegative matrix factorization}.
\newblock \href{https://doi.org/10.1109/TVCG.2013.212}{{\em IEEE Transactions
  on Visualization and Computer Graphics}},
  \href{https://doi.org/10.1109/TVCG.2013.212}{19(12):1992--2001},
  \href{https://doi.org/10.1109/TVCG.2013.212}{2013}.
  \href{https://doi.org/10.1109/TVCG.2013.212}
{doi: {{%
10\hspace{.1pt}\discretionary{.}{%
}{.}\hspace{.4pt}1109\discretionary{/}{%
}{/}TVCG\hspace{.1pt}\discretionary{.}{%
}{.}\hspace{.4pt}2013\hspace{.1pt}\discretionary{.}{%
}{.}\hspace{.4pt}212}}}


\bibitem{collins2022visual}
C.~Collins, A.~Fokkens, A.~Kerren, C.~Weaver, and A.~Chatzimparmpas.
\newblock Visual text analytics: Report from dagstuhl seminar 22191, 2022.

\bibitem{corpus}
B.~Fortuna, M.~Grobelnik, and D.~Mladenić.
\newblock Visualization of text document corpus.
\newblock {\em Informatica (Slovenia)}, 29:497--504, 11 2005.

\bibitem{fryer2022flexible}
Z.~Fryer, V.~Axelrod, B.~Packer, A.~Beutel, J.~Chen, and K.~Webster.
\newblock Flexible text generation for counterfactual fairness probing.
\newblock {\em arXiv preprint arXiv:2206.13757}, 2022.

\bibitem{synthetic_patient_data}
\href{https://doi.org/10.1186/s12874-020-00977-1}{A.~Gonçalves, P.~Ray,
  B.~Soper, J.~Stevens, L.~Coyle, and A.~Sales}.
\newblock \href{https://doi.org/10.1186/s12874-020-00977-1}{Generation and
  evaluation of synthetic patient data}.
\newblock \href{https://doi.org/10.1186/s12874-020-00977-1}{{\em BMC Medical
  Research Methodology}},
  \href{https://doi.org/10.1186/s12874-020-00977-1}{20},
  \href{https://doi.org/10.1186/s12874-020-00977-1}{05 2020}.
  \href{https://doi.org/10.1186/s12874-020-00977-1}
{doi: {{%
10\hspace{.1pt}\discretionary{.}{%
}{.}\hspace{.4pt}1186\discretionary{/}{%
}{/}s12874\discretionary{%
}{-}{-}020\discretionary{%
}{-}{-}00977\discretionary{%
}{-}{-}1}}}


\bibitem{han2001prefixspan}
J.~Han, J.~Pei, B.~Mortazavi-Asl, H.~Pinto, Q.~Chen, U.~Dayal, and M.~Hsu.
\newblock Prefixspan: Mining sequential patterns efficiently by
  prefix-projected pattern growth.
\newblock In {\em Proceedings of the 17th International Conference on Data
  Engineering}, pp. 215--224. IEEE, 2001.

\bibitem{he2022synthetic}
\href{https://arxiv.org/abs/2212.09864}{Z.~He, G.~Blackwood, R.~Panda,
  J.~McAuley, and R.~Feris}.
\newblock \href{https://arxiv.org/abs/2212.09864}{Synthetic pre-training tasks
  for neural machine translation}.
\newblock \href{https://arxiv.org/abs/2212.09864}{{\em arXiv preprint
  arXiv:2212.09864}}, \href{https://arxiv.org/abs/2212.09864}{2022}.

\bibitem{hohman2018visual}
\href{https://doi.org/10.1109/TVCG.2018.2843369}{F.~Hohman, M.~Kahng,
  R.~Pienta, and D.~H. Chau}.
\newblock \href{https://doi.org/10.1109/TVCG.2018.2843369}{Visual analytics in
  deep learning: An interrogative survey for the next frontiers}.
\newblock \href{https://doi.org/10.1109/TVCG.2018.2843369}{{\em IEEE
  Transactions on Visualization and Computer Graphics}},
  \href{https://doi.org/10.1109/TVCG.2018.2843369}{25(8):2674--2693},
  \href{https://doi.org/10.1109/TVCG.2018.2843369}{2018}.
  \href{https://doi.org/10.1109/TVCG.2018.2843369}
{doi: {{%
10\hspace{.1pt}\discretionary{.}{%
}{.}\hspace{.4pt}1109\discretionary{/}{%
}{/}TVCG\hspace{.1pt}\discretionary{.}{%
}{.}\hspace{.4pt}2018\hspace{.1pt}\discretionary{.}{%
}{.}\hspace{.4pt}2843369}}}


\bibitem{spacy2}
M.~Honnibal and I.~Montani.
\newblock {spaCy 2}: Natural language understanding with {B}loom embeddings,
  convolutional neural networks and incremental parsing.
\newblock 2017.

\bibitem{kucher2015text}
K.~Kucher and A.~Kerren.
\newblock Text visualization techniques: Taxonomy, visual survey, and community
  insights.
\newblock In {\em 2015 IEEE Pacific visualization symposium (pacificVis)}, pp.
  117--121. IEEE, 2015.

\bibitem{lara2022evaluation}
H.~Lara and M.~Tiwari.
\newblock Evaluation of synthetic datasets for conversational recommender
  systems.
\newblock {\em arXiv preprint arXiv:2212.08167}, 2022.

\bibitem{tableLens}
\href{https://doi.org/10.1145/191666.191776}{R.~Rao and S.~K. Card}.
\newblock \href{https://doi.org/10.1145/191666.191776}{{The Table Lens}:
  merging graphical and symbolic representations in an interactive focus+
  context visualization for tabular information}.
\newblock \href{https://doi.org/10.1145/191666.191776}{In {\em Proceedings of
  the SIGCHI conference on Human factors in computing systems}},
  \href{https://doi.org/10.1145/191666.191776}{pp. 318--322},
  \href{https://doi.org/10.1145/191666.191776}{1994}.
  \href{https://doi.org/10.1145/191666.191776}
{doi: {{%
10\hspace{.1pt}\discretionary{.}{%
}{.}\hspace{.4pt}1145\discretionary{/}{%
}{/}191666\hspace{.1pt}\discretionary{.}{%
}{.}\hspace{.4pt}191776}}}


\bibitem{reif2019visualizing}
\href{https://proceedings.neurips.cc/paper_files/paper/2019/file/159c1ffe5b61b41b3c4d8f4c2150f6c4-Paper.pdf}{E.~Reif,
  A.~Yuan, M.~Wattenberg, F.~B. Viegas, A.~Coenen, A.~Pearce, and B.~Kim}.
\newblock
  \href{https://proceedings.neurips.cc/paper_files/paper/2019/file/159c1ffe5b61b41b3c4d8f4c2150f6c4-Paper.pdf}{Visualizing
  and measuring the geometry of {BERT}}.
\newblock
  \href{https://proceedings.neurips.cc/paper_files/paper/2019/file/159c1ffe5b61b41b3c4d8f4c2150f6c4-Paper.pdf}{{\em
  Advances in Neural Information Processing Systems}},
  \href{https://proceedings.neurips.cc/paper_files/paper/2019/file/159c1ffe5b61b41b3c4d8f4c2150f6c4-Paper.pdf}{32},
  \href{https://proceedings.neurips.cc/paper_files/paper/2019/file/159c1ffe5b61b41b3c4d8f4c2150f6c4-Paper.pdf}{2019}.

\bibitem{workshop2023bloom}
T.~L. Scao, A.~Fan, C.~Akiki, E.~Pavlick, S.~Ili{\'c}, D.~Hesslow,
  R.~Castagn{\'e}, A.~S. Luccioni, F.~Yvon, M.~Gall{\'e}, et~al.
\newblock Bloom: A 176b-parameter open-access multilingual language model.
\newblock {\em arXiv preprint arXiv:2211.05100}, 2023.

\bibitem{doi:10.1057/palgrave.ivs.9500180}
\href{https://doi.org/10.1057/palgrave.ivs.9500180}{J.~Stasko, C.~Görg, and
  Z.~Liu}.
\newblock \href{https://doi.org/10.1057/palgrave.ivs.9500180}{Jigsaw:
  Supporting investigative analysis through interactive visualization}.
\newblock \href{https://doi.org/10.1057/palgrave.ivs.9500180}{{\em Information
  Visualization}},
  \href{https://doi.org/10.1057/palgrave.ivs.9500180}{7(2):118--132},
  \href{https://doi.org/10.1057/palgrave.ivs.9500180}{2008}.
  \href{https://doi.org/10.1057/palgrave.ivs.9500180}
{doi: {{%
10\hspace{.1pt}\discretionary{.}{%
}{.}\hspace{.4pt}1057\discretionary{/}{%
}{/}palgrave\hspace{.1pt}\discretionary{.}{%
}{.}\hspace{.4pt}ivs\hspace{.1pt}\discretionary{.}{%
}{.}\hspace{.4pt}9500180}}}


\bibitem{strobelt2017lstmvis}
\href{https://doi.org/10.1109/TVCG.2017.2744158}{H.~Strobelt, S.~Gehrmann,
  H.~Pfister, and A.~M. Rush}.
\newblock \href{https://doi.org/10.1109/TVCG.2017.2744158}{{LSTMVis}: A tool
  for visual analysis of hidden state dynamics in recurrent neural networks}.
\newblock \href{https://doi.org/10.1109/TVCG.2017.2744158}{{\em IEEE
  Transactions on Visualization and Computer Graphics}},
  \href{https://doi.org/10.1109/TVCG.2017.2744158}{24(1):667--676},
  \href{https://doi.org/10.1109/TVCG.2017.2744158}{2017}.
  \href{https://doi.org/10.1109/TVCG.2017.2744158}
{doi: {{%
10\hspace{.1pt}\discretionary{.}{%
}{.}\hspace{.4pt}1109\discretionary{/}{%
}{/}TVCG\hspace{.1pt}\discretionary{.}{%
}{.}\hspace{.4pt}2017\hspace{.1pt}\discretionary{.}{%
}{.}\hspace{.4pt}2744158}}}


\bibitem{tang2023does}
\href{https://arxiv.org/abs/2303.04360}{R.~Tang, X.~Han, X.~Jiang, and X.~Hu}.
\newblock \href{https://arxiv.org/abs/2303.04360}{Does synthetic data
  generation of llms help clinical text mining?}
\newblock \href{https://arxiv.org/abs/2303.04360}{{\em arXiv preprint
  arXiv:2303.04360}}, \href{https://arxiv.org/abs/2303.04360}{2023}.

\bibitem{tenney2020language}
\href{https://arxiv.org/abs/2008.05122}{I.~Tenney, J.~Wexler, J.~Bastings,
  T.~Bolukbasi, A.~Coenen, S.~Gehrmann, E.~Jiang, M.~Pushkarna, C.~Radebaugh,
  E.~Reif, and A.~Yuan}.
\newblock \href{https://arxiv.org/abs/2008.05122}{{The Language
  Interpretability Tool}: Extensible, interactive visualizations and analysis
  for {NLP} models}.
\newblock \href{https://arxiv.org/abs/2008.05122}{{\em arXiv preprint
  arXiv:2008.05122}}, \href{https://arxiv.org/abs/2008.05122}{2020}.

\bibitem{vijayakumar2022evaluating}
A.~Vijayakumar.
\newblock Evaluating synthetic code-switched data.
\newblock 2022.

\bibitem{4658133}
\href{https://doi.org/10.1109/TVCG.2008.172}{M.~Wattenberg and F.~B. Viégas}.
\newblock \href{https://doi.org/10.1109/TVCG.2008.172}{The word tree, an
  interactive visual concordance}.
\newblock \href{https://doi.org/10.1109/TVCG.2008.172}{{\em IEEE Transactions
  on Visualization and Computer Graphics}},
  \href{https://doi.org/10.1109/TVCG.2008.172}{14(6):1221--1228},
  \href{https://doi.org/10.1109/TVCG.2008.172}{2008}.
  \href{https://doi.org/10.1109/TVCG.2008.172}
{doi: {{%
10\hspace{.1pt}\discretionary{.}{%
}{.}\hspace{.4pt}1109\discretionary{/}{%
}{/}TVCG\hspace{.1pt}\discretionary{.}{%
}{.}\hspace{.4pt}2008\hspace{.1pt}\discretionary{.}{%
}{.}\hspace{.4pt}172}}}


\bibitem{DBLP:journals/corr/abs-2111-06467}
\href{https://arxiv.org/abs/2111.06467}{A.~Yuan, D.~Ippolito, V.~Nikolaev,
  C.~Callison{-}Burch, A.~Coenen, and S.~Gehrmann}.
\newblock \href{https://arxiv.org/abs/2111.06467}{Synthbio: {A} case study in
  human-ai collaborative curation of text datasets}.
\newblock \href{https://arxiv.org/abs/2111.06467}{{\em arXiv preprint
  arXiv:2111.06467}}, \href{https://arxiv.org/abs/2111.06467}{2021}.

\bibitem{yuan2021survey}
J.~Yuan, C.~Chen, W.~Yang, M.~Liu, J.~Xia, and S.~Liu.
\newblock A survey of visual analytics techniques for machine learning.
\newblock {\em Computational Visual Media}, 7:3--36, 2021.

\end{thebibliography}
